\documentclass[runningheads]{llncs}

\usepackage{eccv}

\usepackage{multirow}

\usepackage{eccvabbrv}

\usepackage{graphicx}
\usepackage{booktabs}

\usepackage[accsupp]{axessibility}  % Improves PDF readability for those with disabilities.

\newcommand{\cavan}[1]{{\color{blue}(cavan: {#1})}} % cavan's comments
\newcommand{\xqx}{\color{black}}
\newcommand{\cam}{\color{black}}

\usepackage{hyperref}

\usepackage{orcidlink}

\begin{document}

% ---------------------------------------------------------------
% TODO REVIEW: Replace with your title
\title{Eliminating Feature Ambiguity for \\ Few-Shot Segmentation} 

% TODO REVIEW: If the paper title is too long for the running head, you can set
% an abbreviated paper title here. If not, comment out.
% \titlerunning{DiscFSS}

% TODO FINAL: Replace with your author list. 
% Include the authors' OCRID for the camera-ready version, if at all possible.
\author{Qianxiong Xu\inst{1}\orcidlink{0000-0001-9175-6783} \and
Guosheng Lin\inst{1}\thanks{Co-corresponding authors}\orcidlink{0000-0002-0329-7458} \and
Chen Change Loy\inst{1}\orcidlink{0000-0001-5345-1591} \and
Cheng Long\inst{1\star}\orcidlink{0000-0001-6806-8405} \and \\
Ziyue Li\inst{2}\orcidlink{0000-0003-4983-9352} \and
Rui Zhao\inst{3}\orcidlink{0000-0001-5874-131X}
}
% TODO FINAL: Replace with an abbreviated list of authors.
\authorrunning{Q. Xu et al.}
% First names are abbreviated in the running head.
% If there are more than two authors, 'et al.' is used.

% TODO FINAL: Replace with your institution list.
\institute{S-Lab, Nanyang Technological University \and University of Cologne \and SenseTime Research\\
\email{\{qianxiong.xu, gslin, ccloy, c.long\}@ntu.edu.sg},\\
\email{zlibn@wiso.uni-koeln.de},
\email{zhaorui@sensetime.com}}

\maketitle

% ========================================
% Abstract
% ========================================
\begin{abstract}
%Recent advances in few-shot segmentation (FSS) have built up pixel-pixel matching between query and support features, using cross attention to activate query foreground (FG) features with the same-class support FG features. 
Recent advancements in few-shot segmentation (FSS) have exploited pixel-by-pixel matching between query and support features, typically based on cross attention, which selectively activate query foreground (FG) features that correspond to the same-class support FG features.
However, due to the large receptive fields in deep layers of the backbone, the extracted query and support FG features are inevitably mingled with background (BG) features, impeding the FG-FG matching in cross attention. Hence, the query FG features are fused with less support FG features, \ie, the support information is not well utilized. This paper presents a novel plug-in termed ambiguity elimination network (AENet), which can be plugged into any existing cross attention-based FSS methods. The main idea is to mine discriminative query FG regions to rectify the ambiguous FG features, increasing the proportion of FG information, so as to suppress the negative impacts of the doped BG features. In this way, the FG-FG matching is naturally enhanced. We plug AENet into
% two baselines CyCTR and SCCAN
{\cam three baselines CyCTR, SCCAN and HDMNet}
for evaluation, and their scores are improved by large margins, \eg,
% the 1-shot performance of SCCAN can be improved by 3.0\% on PASCAL-5$^i$.
{\xqx the 1-shot performance of SCCAN can be improved by 3.0\%+ on both PASCAL-5$^i$ and COCO-20$^i$.
% The source code will be released upon paper acceptance.
{\cam The code is available at \url{https://github.com/Sam1224/AENet}.}
}
% \cavan{How about COCO-20? Promise to release code.}
% \keywords{Few-shot segmentation \and Discriminative prior mask \and Ambiguity elimination} 
{\xqx 
}
% \cavan{keywords should not have the same words taken in the title}

\end{abstract}

% Recent advancements in few-shot segmentation (FSS) have exploited pixel-by-pixel matching between query and support features, typically based on cross attention, which selectively activate query foreground (FG) features that correspond to the same-class support FG features. However, due to the large receptive fields in deep layers of the backbone, the extracted query and support FG features are inevitably mingled with different BG features, impeding the FG-FG matching in cross attention. Hence, the query FG features are fused with less support FG features, \ie, the support information is not well utilized. This paper presents a novel plug-in termed ambiguity elimination network (AENet), which can be plugged into any existing cross attention-based FSS methods. The main idea is to mine discriminative query FG regions to rectify the ambiguous FG features, increasing the proportion of FG information, so as to suppress the negative impacts of the doped BG features. In this way, the FG-FG matching is naturally enhanced. We plug AENet into two baselines CyCTR and SCCAN for evaluation, and their scores are improved by large margins, \eg, the 1-shot performance of SCCAN can be improved by 3.0\%+ on both PASCAL-5$^i$ and COCO-20$^i$. The source code will be released upon paper acceptance.
% \keywords{Discriminative prior mask \and Discriminative query regions \and Feature refinement}

% ========================================
% Introduction
% ========================================
\section{Introduction}
\label{sec:introduction}

% Semantic Segmentation and Few-Shot Segmentation
\if 0
Semantic segmentation is a dense prediction task in computer vision, which involves assigning each image pixel to the correct class label~\cite{garcia2017review,guo2018review,lateef2019survey}.
The rapid evolution of deep learning techniques has led to significant advancements in semantic segmentation \cite{long2015fully,ronneberger2015u,zhao2017pyramid,chen2017deeplab,zhou2022rethinking}. However, they rely heavily on substantial time and human efforts devoted to the meticulous annotation of pixel-wise masks (\eg, ImageNet~\cite{russakovsky2015imagenet}), and the trained model cannot be directly generalized to novel classes (\ie, unseen classes during training).
\fi

Semantic segmentation is a fundamental task within computer vision, entailing the dense assignment of each pixel in an image to an appropriate class label \cite{garcia2017review,guo2018review,lateef2019survey}. 
This process is crucial for understanding the detailed composition of visual scenes. The advent and subsequent evolution of deep learning approaches have led to significant advancements in semantic segmentation \cite{long2015fully,ronneberger2015u,zhao2017pyramid,chen2017deeplab,zhou2022rethinking}. However, most approaches relied heavily on precise pixel-level annotation of training datasets.
% (\eg,
% % ImageNet \cite{russakovsky2015imagenet} 
% {\xqx COCO~\cite{lin2014microsoft}.}
% )
% \cavan{the original dataset doesn't involve mask annotation, use other examples}
Moreover, these models typically exhibit limitations in extending their recognition capabilities to novel, previously unseen classes.

% To address these challenges, the paradigm of few-shot segmentation (FSS) has been introduced to segment query images containing arbitrary classes, with the help of a few support (image, mask) pairs sharing the same class.
{\cam To address these challenges, few-shot segmentation (FSS) has been introduced to segment query images containing arbitrary classes, with the help of a few support (image, mask) pairs sharing the same class.}
During training, FSS models would learn the class-agnostic pattern on some base classes, enabling it to identify query features resembling
{\cam support FG}
% support FG~\footnote{FSS belongs to binary segmentation that involves foreground (FG) and background (BG). Query and support FGs share the same class, while the BGs can be different.}
features and classify them as query FG. 
% Then, the learned pattern can be directly applied to segment arbitrary novel classes.
{\cam Then, such pattern is directly applied to segment novel classes.}

% Summary of existing methods
The effectiveness of FSS heavily relies on the skillful utilization of support samples, based on which the existing methods can be categorized into prototype-based~\cite{zhang2019canet,wang2019panet,li2021adaptive,tian2020prior,fan2022self,bao2023relevant,lang2022learning} and cross attention-based methods~\cite{zhang2019pyramid,wang2020few,zhang2021few,wang2022adaptive,xu2023self,peng2023hierarchical}.
Prototype-based methods typically entail the extraction of support prototypes from support FG features, which are subsequently leveraged to segment the query image through feature comparison~\cite{wang2019panet} or concatenation~\cite{tian2020prior}. Nevertheless, the compression of features into prototypes would lead to potential information loss, as well as the disruption of the spatial structure of objects~\cite{xiong2022doubly}. To address these issues, recent advancements in FSS have embraced cross attention~\cite{vaswani2017attention} to fuse query features with the uncompressed support FG features.

\begin{figure}[tb]
  \centering
  \includegraphics[width=1\linewidth]{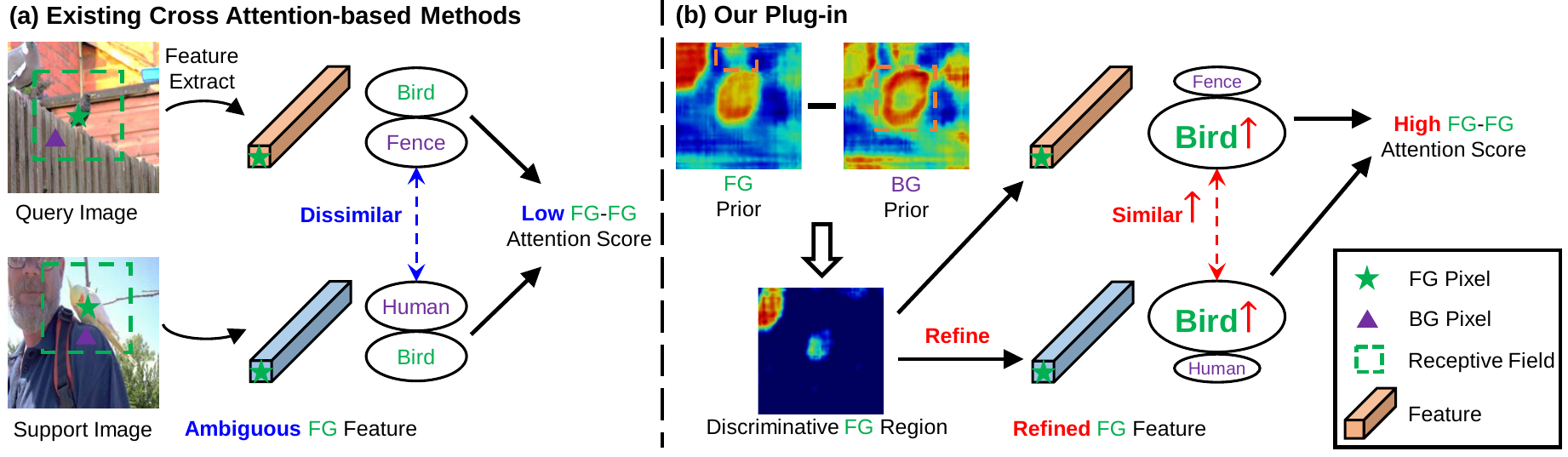}
  \caption{Illustrations of (a) existing methods, and (b) the rationale of our plug-in.
  In (a), due to the large receptive fields during feature extraction, foreground (FG) pixels' features are inevitably fused with background (BG) features (\eg, the bird pixels also contain dissimilar BGs: fence and human), which hinders  FG-FG matching in cross attention.
  In (b), we propose a plug-in to mine discriminative query FG regions, which exclude those regions that are similar to both support FG and BG features, for refining the ambiguous query and support FG features. As a result, the FG parts in the mingled FG features are increased, thus the FG-FG matching is naturally enhanced.
  }
  \label{fig:intro}
  % \vspace{-1em}
\end{figure}

% Limitations
Despite their success, existing cross attention-based methods overlook the \emph{ineffective FG-FG matching} issue raised by \emph{feature ambiguity}. In particular, 
it is a common practice~\cite{shaban2017one,zhang2019canet,wang2019panet} to forward query and support images to a pretrained backbone (\eg, ResNet50~\cite{he2016deep}) to extract their features. However, as illustrated in \cref{fig:intro}(a), deep convolution layers essentially have large receptive fields, so the extracted FG/BG pixels' features are inevitably fused with other BG/FG features, especially for those pixels locating at the boundary area between FG and BG objects.
Some evidences are provided in \cref{fig:intro}(b) in the form of prior masks~\cite{tian2020prior}.
Specifically, cosine similarity is measured between each query feature and support FG features to obtain the FG prior, with each value showing the probability of being a FG pixel. BG prior is obtained in the same way. It can be observed:
(1) There are many wrongly activated query BG regions (in orange rectangle) in FG prior, because they have aggregated nearby FG features, thereby showing high similarity to the support FG;
(2) In BG prior, support BG features can also match with query FG features, because the BG pixels on the border have also been integrated with support FG information.
% As a result, when existing methods~\cite{zhang2021few,xu2023self} conduct cross attention on the ambiguous query and support features:
Take \cref{fig:intro}(a) as an example, where a query FG pixel contains FG (bird) and BG (fence) features, and support FG features include FG (bird) and BG (human). As query and support FG features are doped with different-class BG features, the similarity-based cross attention scores would be relatively smaller, which hinders query FG features from fusing more support FG features, thereby leading to ineffective FSS.

% Solutions
In this paper, we aim to design a plug-in to address the aforementioned issues.
The key idea is to suppress those ambiguous query regions that are similar to both support FG and BG features (doped with some FG information). In this way, the remaining FG regions refer to the most discriminative query FG regions, receiving the least side-effects from the mingled BG features.
Based on this idea, we propose a plug-in named \textbf{ambiguity elimination network (AENet)}, which includes:
(1) \textbf{Prior generator (PG)}:
We incorporate the idea into the learning-agnostic prior mask, and generate a pair of prior masks to figure out the approximate scope of query FG, while highlighting the most discriminative query FG regions, facilitating fast convergence~\cite{tian2020prior}. The visualizations (in \cref{fig:intro}(b) and \cref{fig:qualitative_prior}(b)) can directly show the effectiveness of the idea;
(2) \textbf{Ambiguity eliminator (AE)}:
The idea is further employed to rectify the query and support FG features, so as to naturally improve the FG-FG matching in cross attention, \ie, to better utilize the support information.
As shown in \cref{fig:intro}(b), the features of the discriminative query FG regions are fused with the query and support features for refinement. For a query FG pixel, its features would consequently contain more FG information, so the side-effects of the mingled BG features can be suppressed.
In this paper, we validate the effectiveness of AENet on 
% two cross attention-based FSS baselines, namely, CyCTR~\cite{zhang2021few} and SCCAN~\cite{xu2023self}.
{\cam three cross attention-based FSS baselines, CyCTR~\cite{zhang2021few}, SCCAN~\cite{xu2023self} and HDMNet~\cite{peng2023hierarchical}.}
%
% Contributions
\if 0
To sum up, our contributions are summarized as follows:
\begin{itemize}
    \item To the best of our knowledge, we are the first to identify the negative impacts of feature ambiguity on cross attention-based FSS methods. 
    \item We propose a simple yet effective idea to obtain the discriminative query FG regions. Then, we propose the plug-in network AENet to enhance the FG-FG matching for existing cross attention-based FSS methods.
    \item Extensive experiments are conducted on two public benchmark PASCAL-5$^i$ and COCO-20$^i$ to validate the effectiveness of AENet. Notably, AENet is plugged into CyCTR and SCCAN, and can improve their performance by large margins, \eg, the performance of SCCAN can be boosted from 66.8\% to 69.8\% on PASCAL-5$^i$, under 1-shot setting. \cavan{performing experiments and getting some good results cannot be considered as a contribution.} 
\end{itemize}
\fi

To our knowledge, we are the first to identify the negative impacts of feature ambiguity on
% cross attention-based FSS methods.
{\cam FSS.}
We propose a simple yet effective idea to obtain the discriminative query FG regions. Then, we propose the plug-in network AENet to enhance the FG-FG matching for existing cross attention-based FSS methods. Extensive experiments are conducted on two public benchmarks PASCAL-5$^i$ and COCO-20$^i$ to validate the effectiveness of AENet. Notably, AENet is plugged into
% CyCTR and SCCAN,
{\cam CyCTR, SCCAN and HDMNet,}
and can improve their performance by large margins, \eg,
% the performance of SCCAN can be boosted from 66.8\% to 69.8\% on PASCAL-5$^i$, under 1-shot setting.
{\xqx 
the 1-shot performance of SCCAN can be boosted from 66.8\% to 69.8\%, and 46.3\% to 49.4\% on PASCAL-5$^i$ and COCO-20$^i$, respectively.
}
% \cavan{How about COCO-20?}

% ========================================
% Related Work
% ========================================
\section{Related Work}
\label{sec:related_work}

% \cavan{Missing ref? Distilling Self-Supervised Vision Transformers for Weakly-Supervised Few-Shot Classification \& Segmentation}
% {\xqx I just added this paper to the special setting group. (1) It leverages a more advanced pretrained backbone (DINO ViT), compared to all other baselines; (2) It belongs to semi-supervised learning, while all other methods are fully-supervised.}

\noindent
\textbf{Few-shot segmentation.}
Different from semantic segmentation models that are trained and tested on the same set of classes, FSS is designed to segment arbitrary classes with the help of a few labelled samples.
FSS methods in recent literature can be broadly categorized into several groups.

Prototype-based methods~\cite{shaban2017one,wang2019panet,tian2020prior,li2021adaptive,lang2022beyond,liu2022learning,zhang2021self,fan2022self,okazawa2022interclass,wang2023adaptive,wang2024adaptive} represent a prominent category within the realm of FSS, wherein support FG information is compressed into single or multiple prototypes, facilitating the segmentation of query images. These methodologies typically leverage techniques such as cosine similarity or feature concatenation to perform segmentation.
In particular, OSLSM~\cite{shaban2017one} makes pioneering contributions to the field by introducing the concept of FSS, laying the groundwork for subsequent research endeavors.
PFENet~\cite{tian2020prior} firstly derives a learning-agnostic query prior mask from the high-level query and support features to help coarsely locate the query FG objects and improve the convergence speed.
Each value in the prior mask represents the feature similarity between the current query pixel to the support FG pixels.
However, the existing method is heavily affected by noises~\cite{xu2023self}, and there are many wrong responses on query BG areas. In this paper, we propose a prior generator (PG) to highlight the most discriminative query regions that can accurately locate query FG.

To prevent from the information loss and the structure disruption raised by prototypes, attention-based methods \cite{zhang2021few,zhang2019pyramid,wang2020few,hu2019attention,xie2021scale,jiao2022mask,xu2023self} build pixel-pixel matching between query and support features, and expect to learn similarity-based cross attentions to activate query FG features with the same-class support FG features.
PGNet~\cite{zhang2019pyramid} builds up two graphs on the query and support features, then incorporate the attention mechanism with multi-scale graph reasoning to perform segmentation.
Besides, CyCTR~\cite{zhang2021few} devises a cycle-consistent attention to suppress the side-effects of harmful support features.
Recently, SCCAN~\cite{xu2023self} indicates the mismatch problem of query BG features to the support FG features when conducting cross attention, and propose a self-calibrated cross attention block to align query BG features with appropriate BG features for effective segmentation.
Another lane of methods~\cite{hong2022cost,xiong2022doubly,shi2022dense,liu2023fecanet,park2023task} regard the problem as semantic correspondence and use memory-expensive 4D attentions for the matching.
% Unfortunately, they all ignore the fact that the features of query and support FG pixels would be inevitably fused with different BG features, which is raised by the large receptive fields in the backbone. When performing cross attention, these different BG information would impede the matching between query and support FG.
{\cam Unfortunately, it would be ineffective for these methods to perform query-support FG-FG matching (via cross attention), because the extracted query and support FG features are likely to be mingled with dissimilar BG features, impeding the effective utilization of support information.}
To address it, we propose an ambiguity eliminator (AE) to purify the FG features and make the FG-FG matching effective.

Moreover, many methods extend the standard setting by using base class predictions~\cite{lang2022learning,sun2022singular,iqbal2022msanet,peng2023hierarchical,zhu2024addressing}, mining knowledge from unlabelled data~\cite{bao2023relevant} or more advanced pretrained backbone~\cite{kang2023distilling}, or introducing extra textual information~\cite{yang2023mianet,zhu2023llafs}. To have better comparisons with recent advances, we have conducted experiments under both standard and BAM's~\cite{lang2022learning} settings in \cref{sec:component_ablation}.

% ========================================
% Problem Definition
% ========================================
\section{Problem Definition}
\label{sec:problem_definition}

Let $\mathcal{D}_{\text{train}}$ and $\mathcal{D}_{\text{test}}$ represent the training and testing sets, respectively, in the context of FSS. $\mathcal{D}_{\text{train}}$ encompasses a collection of base classes $\mathcal{C}_{\text{base}}$, while $\mathcal{D}_{\text{test}}$ encompasses another distinct set of novel classes $\mathcal{C}_{\text{novel}}$. FSS addresses a scenario where the sets of classes $\mathcal{C}_{\text{base}}$ and $\mathcal{C}_{\text{novel}}$ are disjoint, formally denoted as $\mathcal{C}_{\text{base}} \cap \mathcal{C}_{\text{novel}} = \emptyset$.
Both $\mathcal{D}_{\text{train}}$ and $\mathcal{D}_{\text{test}}$ are composed of numerous episodes, which serve as the fundamental units of episodic training. In the context of a $k$-shot setting, each episode comprises a support set $\mathcal{S} = \{I_S^n,M_S^n\}_{n=1}^{k}$ and a query set $\mathcal{Q} = \{I_Q,M_Q\}$ specific to a particular class $c$. Here, $I_S^n$ and $M_S^n$ denote the $n$-th support image and its corresponding annotated binary mask, while $I_Q$ and $M_Q$ represent the query image and its associated mask.
During the training phase, the model learns to segment the query image $I_Q$ with guidance from the support set $\mathcal{S}$, focusing on classes from $\mathcal{C}_{\text{base}}$. Subsequently, this learned segmentation pattern is applied to $\mathcal{C}_{\text{novel}}$ during the testing phase.

% ========================================
% Methodology
% ========================================
\section{Methodology}
\label{sec:methodology}

\begin{figure}[tb]
  \centering
  \includegraphics[width=0.95\linewidth]{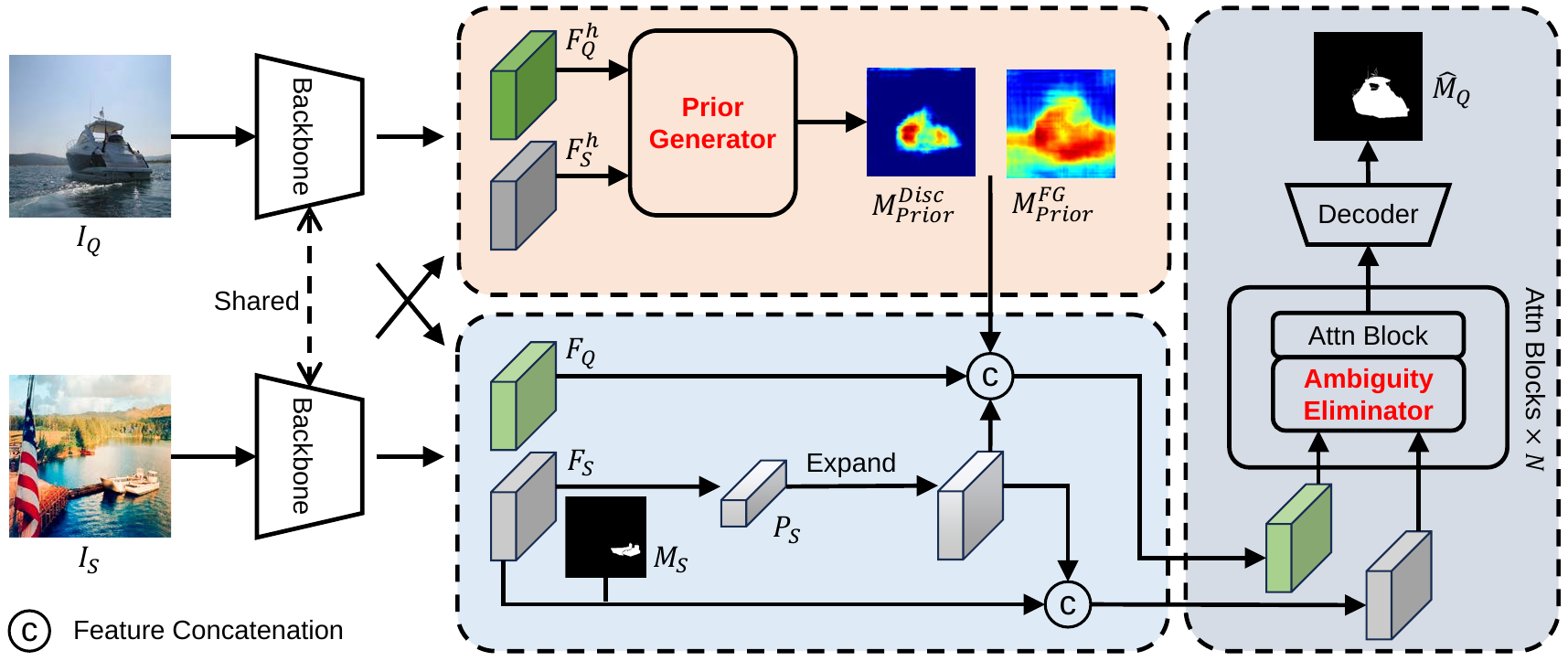}
  \caption{Overview of ambiguity elimination network (AENet), which is designed for existing cross attention-based methods such as SCCAN~\cite{xu2023self}. Specifically, AENet includes prior generator (PG) and ambiguity eliminator (AE). (1) Learning-agnostic PG helps to accurately locate the query FG regions;
  % (2) AE mines the most discriminative query FG regions that are less affected by the mingled BG features, and then rectifies the query and support FG features to improve the FG-FG matching in cross attention.
  (2) AE mines the most discriminative query FG regions, and then rectifies the query and support FG features to improve the FG-FG matching in cross attention.
  }
  \label{fig:overview}
  % \vspace{-1em}
\end{figure}

The standard FSS pipeline is described as follows:
The query image $I_Q$ and the support image $I_S$ are forwarded to a frozen ImageNet-pretrained~\cite{russakovsky2015imagenet} backbone (\eg, VGG16~\cite{simonyan2014very} or ResNet50~\cite{he2016deep}) for extracting their mid-level features $\{F_Q, F_S\}$ (from blocks 2 and 3) and high-level features $\{F_Q^h, F_S^h\}$ (from block 4). As explained in existing methods~\cite{zhang2019canet,tian2020prior}, FSS models would witness severe overfitting phenomena if segmentation is carried out with the high-level features. In spite of this, PFENet~\cite{tian2020prior} demonstrates that learning-agnostic prior masks can be derived from high-level features to coarsely locate the query FG objects and improve the convergence speed. Therefore, almost all recent advances~\cite{lang2022learning,peng2023hierarchical,xu2023self} follow the same guideline to take the mid-level features for segmentation, and use the high-level features for generating prior masks.
Then, support FG features are obtained with $F_S$ and support mask $M_S$, which are fused with query features, either in a prototypical way or a cross attention manner, to activate those query FG features that share the same class.
Finally, the enhanced query features are processed by a decoder to obtain the predicted query mask $\hat{M}_Q$.

To mitigate the \emph{ineffective FG-FG matching} issue and \emph{feature ambiguity} issue (as described in \cref{sec:introduction}), we present the overview of our plug-in ambiguity elimination network (AENet) in \cref{fig:overview}, whose main components consist of the prior generator (PG) module (\cref{sec:dpm}) and the ambiguity eliminator (AE) module (\cref{sec:ae}). AENet can be plugged into any existing cross attention-based FSS baselines such as
% CyCTR~\cite{zhang2021few} and SCCAN~\cite{xu2023self}.
{\cam CyCTR~\cite{zhang2021few}, SCCAN~\cite{xu2023self} and HDMNet~\cite{peng2023hierarchical}.}
Take SCCAN as an example, its pseudo mask aggregation (PMA) module is replaced with our proposed PG, and we insert one AE before each of its self-calibrated cross attention (SCCA) block, with other parts remaining untouched. Particularly, the learning-agnostic PG can not only approximately locate the query FG, but also extract the discriminative query FG regions. In this way, the FSS model can obtain some hints with no additional learnable parameters, and its converge speed is well improved. Besides, the AE module aims at using the rectified discriminative query FG regions to refine the query FG and support FG features, so as to suppress the negative impacts of their doped BG features. As a result, the FG-FG matching in cross attention is enhanced, leading to more effective FSS.

\subsection{Prior Generator (PG)}
\label{sec:dpm}

\begin{figure}[tb]
  \centering
  \includegraphics[width=1\linewidth]{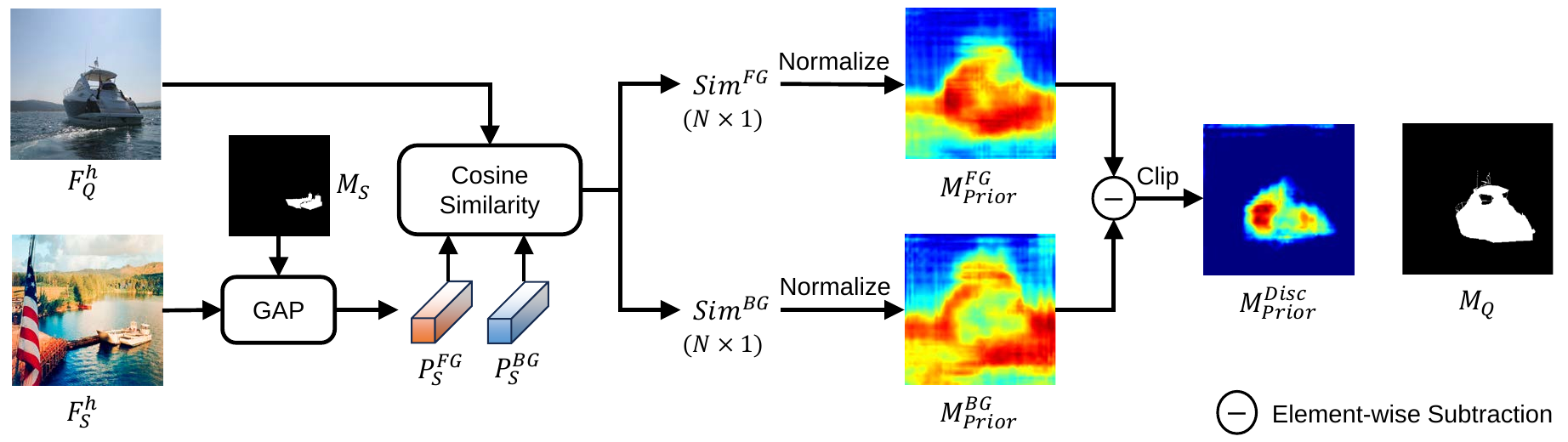}
  \caption{Details of the learning-agnostic prior generator (PG) module.
  % The query mask $M_Q$ is included only for comparison purpose.
  The discriminative prior mask can suppress those ambiguous query regions that are similar to both support FG and BG features.
  }
  \label{fig:dpm}
  % \vspace{-1em}
\end{figure}

Since PFENet~\cite{tian2020prior}, learning-agnostic prior masks are widely adopted in FSS models~\cite{zhang2021few,xu2023self,peng2023hierarchical}, to coarsely locate the query FG objects and improve the convergence speed. For each query pixel, most of the existing methods~\cite{tian2020prior,liu2022dynamic,luo2021pfenet++,zhang2021few} take its high-level features, calculate its cosine similarity scores with each support FG feature, and take the normalized maximum score to show how likely this query pixel belongs to FG class. Later, SCCAN~\cite{xu2023self} indicates the operation of taking a single maximum score is heavily affected by noises, and they propose to utilize all pairwise similarities for generating more robust prior masks.

Nevertheless, they overlook the fact that the high-level FG and BG features are inevitably fused with BG and FG features (\ie, feature ambiguity), which means query-support FG-FG may not consistently have large similarity scores, while query-support BG-FG can also mistakenly have high similarity. Hence, the generated prior masks would be less effective. We support our claim by providing some
% direct evidences
{\cam examples}
in \cref{fig:dpm}, where $M_{Prior}^{FG}$ and $M_{Prior}^{BG}$ are the query prior masks calculated with support FG and BG features, respectively. We could observe that (1) there are many BG regions wrongly activated in $M_{Prior}^{FG}$, and (2) support BG can also have high responses on query FG in $M_{Prior}^{BG}$.

To tackle the aforementioned issue, we propose to rectify and extract the most discriminative query FG regions via a simple subtraction operation between $M_{Prior}^{FG}$ and $M_{Prior}^{BG}$, whose essence is removing those ambiguous query regions that show high similarity to both support FG and BG features. As shown in \cref{fig:dpm}, the discriminative prior mask $M_{Prior}^{Disc}$ can remove the unexpected areas, suppressing the side-effects of the doped FG information in support BG features.

The details of PG are presented in \cref{fig:dpm}, and we formally describe it as follows. Following existing methods, we take the high-level query features $F_Q^h$ and support features $F_S^h$, as well as the downsampled support mask $M_S$ as inputs. First of all, the support FG and BG prototypes $P_S^{FG}$ and $P_S^{BG}$ are obtained via:
\begin{equation}\label{eq:proto_extraction}
    P_S^{FG} = GAP(F_S^h, M_S), P_S^{BG} = GAP(F_S^h, 1 - M_S)
\end{equation}
where $GAP(\cdot)$ denotes the global average pooling (GAP) operation.
Then, two cosine similarity scores $Sim^{FG} \in [-1, 1]$ and $Sim^{BG} \in [-1, 1]$ are calculated between the flattened query features and the support prototypes. They are normalized and reshaped to obtain the prior masks $M_{Prior}^{FG} \in [0, 1]$ and $M_{Prior}^{BG} \in [0, 1]$, showing the query regions that are similar to support FG and BG features, respectively.
Notably, our memory cost is $HW \times 1$ for each calculation, while that of existing methods~\cite{tian2020prior,xu2023self} is $HW \times HW$.
\begin{align}
    Sim^{*} &= Cosine(F_Q^h, P_S^{*}) \;
    \\
    M_{Prior}^{*} &= Norm(Sim^{*}) \label{eq:m_prior_fg_bg}
\end{align}
where superscript $* \in \{FG, BG\}$, $Cosine(\cdot)$ means cosine similarity, $Norm(\cdot)$ is the min-max scaler to normalize the values into $[0, 1]$.
Next, we perform a clipped subtraction to rectify and obtain the discriminative prior mask $M_{Prior}^{Disc}$.
% \begin{equation}\label{eq:disc_prior}
%     M_{Prior}^{Disc} = Clip(M_{Prior}^{FG} - M_{Prior}^{BG})
% \end{equation}
\begin{equation}\label{eq:disc_prior}
    M_{Prior}^{Disc} = ReLU(M_{Prior}^{FG} - M_{Prior}^{BG})
\end{equation}
where
% $Clip(\cdot)$
{\cam $ReLU(\cdot)$}
is an operator to set the negative values as zeros. The negative values correspond to those query regions that are similar to support BG features, which are not helpful in FSS.
Finally, we concatenate $M_{Prior}^{FG}$ and $M_{Prior}^{Disc}$ as the final prior masks, providing both the coarse location of query FG objects and the most discriminative query FG regions.

\subsection{Ambiguity Eliminator (AE)}
\label{sec:ae}

\begin{figure}[tb]
  \centering
  \includegraphics[width=0.82\linewidth]{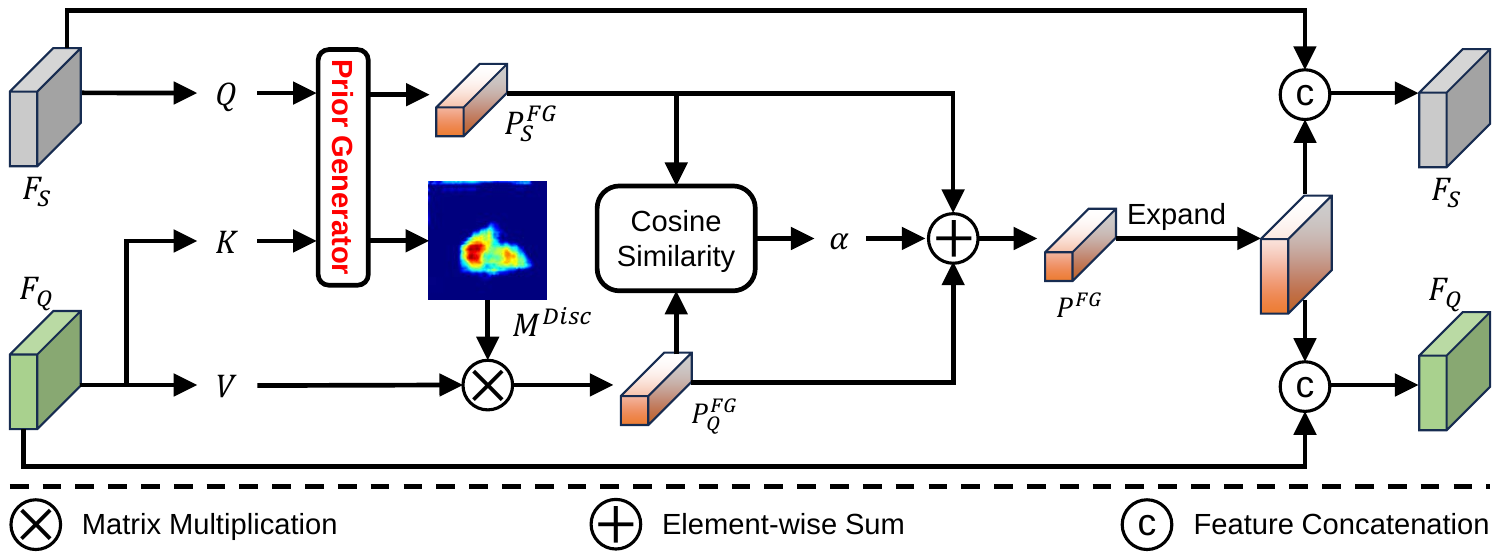}
  \caption{Details of ambiguity eliminator (AE) module. AE mines discriminative query FG features to rectify the query and support features.
  % AE is wrapped by a Transformer block.
  % \ie, the input features would be sequentially processed by layer normalization (LN), AE, LN, and feed-forward network (FFN), with two skip connections.
  }
  \label{fig:ae}
  % \vspace{-1em}
\end{figure}

To tackle the \emph{ineffective FG-FG matching} raised by \emph{feature ambiguity}, we further design the AE module for existing cross attention-based FSS methods~\cite{zhang2021few,xu2023self,peng2023hierarchical}.
As verified in \cref{fig:dpm}, discriminative query FG regions can be easily and consistently extracted, whose features are the most discriminative ones that are less affected by \emph{feature ambiguity}. Therefore, we use them to rectify the query and support FG features, so the proportion of pure FG information in a FG pixel's mingled features can be naturally increased. Therefore, the FG-FG matching between query and support FG features is enhanced, and the query FG pixels can aggregate more support FG information.

As illustrated in \cref{fig:ae}, the mid-level query features $F_Q$  and support features $F_S$ are taken as the inputs. $F_Q$ is processed by two linear layers to obtain the projections $K$ and $V$, while $F_S$ is projected to $Q$. Then, the $Q$ and $K$ are forwarded to the PG module to obtain the support FG prototype $P_S^{FG}$ and the discriminative prior mask $M^{Disc}$, which is supervised by an auxiliary binary cross entropy (BCE) loss $\mathcal{L}_{aux}$. The procedure can be written as:
\begin{align}
    Q = Linear(F_S), K &= Linear(F_Q), V = Linear(F_Q) \;
    \\
    P_S^{FG}, M^{Disc} &= PG(Q, K) \; \label{eq:mid_dpm}
    \\
    \mathcal{L}_{aux} &= BCE(M^{Disc}, M_Q) \label{eq:aux_loss}
\end{align}
where $PG(\cdot)$
% means processing the inputs through 
{\cam includes}
\cref{eq:proto_extraction} to \cref{eq:disc_prior}, $\mathcal{L}_{aux}$ is utilized during training, $M_Q$ is the labelled query mask.
Then, a matrix multiplication is performed to extract and aggregate the discriminative query features into a prototype $P_Q^{FG}$. Next, the cosine similarity $\alpha$ between $P_S^{FG}$ and $P_Q^{FG}$ is measured, and re-scaled to $[0, 1]$. $\alpha$ is utilized to weighted fuse the prototypes to obtain $P^{FG}$, containing the support FG features and the most discriminative query FG features. After that, $P^{FG}$ is expanded and concatenated with the input query and support features $F_Q$ and $F_S$, and processed by a linear layer for feature refinement.
\begin{align}
    P_Q^{FG} &= Softmax(M^{Disc}) \otimes V \;
    \\
    \alpha &= (Cosine(P_S^{FG}, P_Q^{FG}) + 1) / 2 \;
    \\
    P^{FG} &= \alpha \cdot P_S^{FG} + (1 - \alpha) \cdot P_Q^{FG} \;
    \\
    F_* &= Linear(F_* || P^{FG})
\end{align}
where $\otimes$ is the matrix multiplication operation, subscript $* \in \{Q, S\}$, $||$ denotes feature concatenation.
Finally, we wrap AE with a Transformer block~\cite{vaswani2017attention}, and the refined query and support features are forwarded to existing cross attention blocks for feature matching and enhancement.

\subsection{Overall Loss}
\label{sec:overall_loss}

The loss function consists of a main loss and an auxiliary loss. The former refers to the loss functions employed in the original baseline, while the auxiliary loss corresponds to \cref{eq:aux_loss}. Take SCCAN as an example, its loss function is:
\begin{align}\label{eq:overall_loss}
    \mathcal{L} &= \mathcal{L}_{main} + \lambda \cdot \mathcal{L}_{aux} \;
    \\
    &= Dice(\hat{M}_Q, M_Q) + \lambda \cdot (\frac{1}{N}\sum_{i=1}^{N}BCE(M^{Disc}_i, M_Q))
\end{align}
where $Dice(\cdot)$ represents dice loss, $\hat{M}_Q$ is the final prediction, $\lambda$ is a hyperparameter, $N$ is the number of attention blocks (\eg, SCCAN has 8 SCCA blocks), $M_i^{Disc}$ is the $i$-th discriminative mask obtained from the $i$-th AE module.

% ========================================
% Experiments
% ========================================
\section{Experiments}
\label{sec:experiments}

\subsection{Experiment Setup}
\label{sec:experimental_settings}

\noindent
\textbf{Datasets.}
We assess the performance of our methodology on two widely used benchmark datasets, including PASCAL-5$^i$~\cite{shaban2017one} and COCO-20$^i$~\cite{nguyen2019feature}. PASCAL-5$^i$ comprises 20 distinct classes and is derived from PASCAL VOC 2012~\cite{everingham2010pascal}, augmented with additional annotations from \cite{hariharan2014simultaneous}. Conversely, COCO-20$^i$ is constructed from MSCOCO~\cite{lin2014microsoft}, presenting a more rigorous challenge with its 80 classes.
Both PASCAL-5$^i$ and COCO-20$^i$ are partitioned into four folds for cross-validation purposes, with each fold containing either 5 (for PASCAL-5$^i$) or 20 (for COCO-20$^i$) classes. Within each fold, the training set encompasses the union of the other three folds, while the fold itself is reserved for testing. Furthermore, during testing, we randomly sample 1,000 episodes from PASCAL-5$^i$ and 4,000 episodes from COCO-20$^i$, ensuring a comprehensive evaluation of our method's efficacy across diverse scenarios.

\noindent
\textbf{Evaluation metrics.}
It is a common practice to use mean Intersection over Union (mIoU) and foreground-background IoU (FB-IoU) as the evaluation metrics~\cite{tian2020prior,xu2023self,peng2023hierarchical}. Specifically, mIoU computes the average IoU scores across all FG classes within a fold, providing a consolidated measure of segmentation accuracy. Instead, FB-IoU treats all FG classes as a single FG class for measurement.

\noindent
\textbf{Implementation details.}
To validate the effectiveness of AENet, we plug it into
% two
{\cam three}
cross attention-based baselines:
% CyCTR~\cite{zhang2021few} and SCCAN~\cite{xu2023self}.
{\cam CyCTR~\cite{zhang2021few}, SCCAN~\cite{xu2023self} and HDMNet~\cite{peng2023hierarchical}.}
In detail, their prior mask generation modules are replaced by PG (\cref{sec:dpm}), and we insert an AE module (\cref{sec:ae}) before each of their cross attention blocks.
All experiments are conducted with 4 NVIDIA V100 GPUs with 16GB onboard memory.
For both datasets,
we
% crop the images to 473$\times$473, and 
adopt the same augmentation strategy as \cite{tian2020prior,lang2022learning}. The batch size is fixed as 8, and we follow the selected baselines to set optimizer-related parameters.
In this paper, we perform evaluation under both standard~\cite{tian2020prior} and BAM's~\cite{lang2022learning} settings. For the former, we adopt VGG16~\cite{simonyan2014very} and ResNet50~\cite{he2016deep} pretrained on ImageNet~\cite{russakovsky2015imagenet} as backbones, while for the latter, they are further fine-tuned for segmenting
% base classes
$\mathcal{C}_{base}$, which is more appropriate for FSS.
Besides, we follow the best hyperparameter settings as the baselines (\eg, number of attention blocks), while for AENet-related parameters, the weight $\lambda$ (\cref{eq:overall_loss}) is set to 1, and is further studied in \cref{sec:weight_lambda}.

\subsection{Quantitative Comparisons with State of the Arts}
\label{sec:quantitative_sota}

\begin{table}[tb]
  \caption{Performance comparisons on PASCAL-5$^i$ in terms of mIoU and FB-IoU. ``5$^i$'' shows the mIoU scores of 5 novel classes in fold $i$, ``Mean'' is the averaged mIoU score from all folds. The best results are highlighted in \textbf{bold}.}
  \label{tab:quantitative_pascal}
  \centering
  \resizebox{1\linewidth}{!}{
    \begin{tabular}{@{}cl|cccccc|cccccc@{}}
    \toprule
    \multirow{2}[0]{*}{Backbone} & \multirow{2}[0]{*}{Method} & \multicolumn{6}{c|}{1-shot}                    & \multicolumn{6}{c}{5-shot} \\
          &       & 5$^0$ & 5$^1$ & 5$^2$ & 5$^3$ & Mean  & FB-IoU & 5$^0$ & 5$^1$ & 5$^2$ & 5$^3$ & Mean  & FB-IoU \\
    \midrule
    \multirow{8}[0]{*}{VGG16} & PFENet (TPAMI'20)~\cite{tian2020prior} & 56.9  & 68.2  & 54.4  & 52.5  & 58.0  & 72.0  & 59.0  & 69.1  & 54.8  & 52.9  & 59.0  & 72.3 \\
          & DACM (ECCV'22)~\cite{xiong2022doubly} & 61.8  & 67.8  & 61.4  & 56.3  & 61.8  & 75.5  & 66.1  & 70.6  & 65.8  & 60.2  & 65.7  & 77.8 \\
          & FECANet (TMM'23)~\cite{liu2023fecanet} & \textbf{66.5}  & 68.9  & 63.6  & 58.3  & 64.3  & 76.2  & 68.6  & 70.8  & 66.7  & 60.7  & 66.7  & 77.6 \\
          & BAM (CVPR'22)~\cite{lang2022learning} & 63.2  & 70.8  & 66.1  & 57.5  & 64.4  & 77.3  & 67.4  & 73.1  & 70.6  & 64.0  & 68.8  & 81.1 \\
          \cmidrule{2-14}
          & HDMNet (CVPR'23)~\cite{peng2023hierarchical} & 64.8  & 71.4  & 67.7  & 56.4  & 65.1  & -     & 68.1  & 73.1  & 71.8  & 64.0  & 69.3  & - \\
          & HDMNet + \textbf{AENet} & 64.7 & 73.2 & 66.4 & \textbf{60.1} & 66.1$\mathbf{_{1.0\uparrow}}$ & 78.1 & 69.1 & 74.8 & 69.4 & \textbf{65.1} & 69.6$\mathbf{_{0.3\uparrow}}$ & \textbf{82.0} \\
          & SCCAN (ICCV'23)~\cite{xu2023self} & 63.3 & 70.8 & 66.6 & 58.2 & 64.7 & 77.2 & 67.2 & 72.3 & 70.5 & 63.8 & 68.4 & 79.1 \\
          & SCCAN + \textbf{AENet} & 66.3 & \textbf{73.3} & \textbf{68.5} & 58.4 & \textbf{66.6$\mathbf{_{1.9\uparrow}}$} & \textbf{79.0} & \textbf{70.8} & \textbf{75.1} & \textbf{72.2} & 64.2 & \textbf{70.6$\mathbf{_{2.2\uparrow}}$} & 81.8 \\
    \midrule
    \multirow{13}[0]{*}{ResNet50} & PFENet (TPAMI'20) & 61.7  & 69.5  & 55.4  & 56.3  & 60.8  & 73.3  & 63.1  & 70.7  & 55.8  & 57.9  & 61.9  & 73.9 \\
          & DACM (ECCV'22)~\cite{xiong2022doubly} & 66.5  & 72.6  & 62.2  & 61.3  & 65.7  & 77.8  & 72.4  & 73.7  & 69.1  & 68.4  & 70.9  & 81.3 \\
          & ABCNet (CVPR'23)~\cite{wang2023rethinking} & 68.8  & 73.4  & 62.3  & 59.5  & 66.0  & 76.0  & 71.7  & 74.2  & 65.4  & 67.0  & 69.6  & 80.0 \\
          & RiFeNet (AAAI'24)~\cite{bao2023relevant} & 68.4  & 73.5  & 67.1  & 59.4  & 67.1  & -     & 70.0  & 74.7  & 69.4  & 64.2  & 69.6  & - \\
          & FECANet (TMM'23)~\cite{liu2023fecanet} & 69.2  & 72.3  & 62.4  & \textbf{65.7}  & 67.4  & 78.7  & 72.9  & 74.0  & 65.2  & 67.8  & 70.0  & 80.7 \\
          & BAM (CVPR'22)~\cite{lang2022learning} & 69.0  & 73.6  & 67.6  & 61.1  & 67.8  & 79.7  & 70.6  & 75.1  & 70.8  & 67.2  & 70.9  & 82.2 \\
          & MIANet (CVPR'23)~\cite{yang2023mianet} & 68.5  & 75.8  & 67.5  & 63.2  & 68.7  & 79.5  & 70.2  & 77.4  & 70.0  & 68.8  & 71.6  & 82.2 \\
          \cmidrule{2-14}
          & CyCTR (NIPS'21)~\cite{zhang2021few} & 67.8 & 72.8 & 58.0 & 58.0 & 64.2 & - & 71.1 & 73.2 & 60.5 & 57.5 & 65.6 & - \\
          & CyCTR + \textbf{AENet} & 70.4 & 74.4 & \textbf{69.4} & 61.6 & 69.0$\mathbf{_{4.8\uparrow}}$ & 80.2 & 73.0 & 76.2 & 74.1 & 67.0 & 72.6$\mathbf{_{7.0\uparrow}}$ & 83.0 \\
          & SCCAN (ICCV'23)~\cite{xu2023self} & 68.3 & 72.5 & 66.8 & 59.8 & 66.8 & 77.7 & 72.3 & 74.1 & 69.1 & 65.6 & 70.3 & 81.8 \\
          & SCCAN + \textbf{AENet} & \textbf{72.2} & 75.5 & 68.5 & 63.1 & 69.8$\mathbf{_{3.0\uparrow}}$ & 80.8 & \textbf{74.2} & 76.5 & \textbf{74.8} & 70.6 & 74.1$\mathbf{_{3.8\uparrow}}$ & \textbf{84.5} \\
          & HDMNet (CVPR'23)~\cite{peng2023hierarchical} & 71.0  & 75.4  & 68.9  & 62.1  & 69.4  & -     & 71.3  & 76.2  & 71.3  & 68.5  & 71.8  & - \\
          & HDMNet + \textbf{AENet} & 71.3 & \textbf{75.9} & 68.6 & 65.4 & \textbf{70.3$\mathbf{_{0.9\uparrow}}$} & \textbf{81.2} & 73.9 & \textbf{77.8} & 73.3 & \textbf{72.0} & \textbf{74.2$\mathbf{_{2.4\uparrow}}$} & \textbf{84.5} \\
    \bottomrule
    \end{tabular}
  }
  % \vspace{-1em}
\end{table}

\begin{table}[tb]
  \caption{Performance comparisons on COCO-20$^i$ in terms of mIoU and FB-IoU. ``20$^i$'' shows the mIoU scores of 20 novel classes in fold $i$, ``Mean'' is the averaged mIoU score from all folds. $^*$ means the reproduced results. The best results are highlighted in \textbf{bold}.}
  \label{tab:quantitative_coco}
  \centering
  \resizebox{1\linewidth}{!}{
    \begin{tabular}{@{}cl|cccccc|cccccc@{}}
    \toprule
    \multirow{2}[0]{*}{Backbone} & \multirow{2}[0]{*}{Method} & \multicolumn{6}{c|}{1-shot}                    & \multicolumn{6}{c}{5-shot} \\
          &       & 5$^0$ & 5$^1$ & 5$^2$ & 5$^3$ & Mean  & FB-IoU & 5$^0$ & 5$^1$ & 5$^2$ & 5$^3$ & Mean  & FB-IoU \\
    \midrule
    \multirow{7}[0]{*}{VGG16} & PFENet (TPAMI'20)~\cite{tian2020prior} & 35.4  & 38.1  & 36.8  & 34.7  & 36.3  & 63.3  & 38.2  & 42.5  & 41.8  & 38.9  & 40.4  & 65.0 \\
          & FECANet (TMM'23)~\cite{liu2023fecanet} & 34.1  & 37.5  & 35.8  & 34.1  & 35.4  & 65.5  & 39.7  & 43.6  & 42.9  & 39.7  & 41.5  & 67.7 \\
          & BAM (CVPR'22)~\cite{lang2022learning} & 39.0  & 47.0  & 46.4  & 41.6  & 43.5  & 71.1 & 47.0  & 52.6  & 48.6  & 49.1  & 49.3  & 73.3 \\
          \cmidrule{2-14}
          & HDMNet$^*$ (CVPR'23)~\cite{peng2023hierarchical} & \textbf{42.6} & 49.6 & 46.7 & 44.7 & 45.9 & 71.4 & 48.0 & 55.8 & 54.7 & 50.5 & 52.2 & 74.8 \\
          & HDMNet + \textbf{AENet} & 41.9 & 49.5 & \textbf{49.2} & \textbf{45.0} & \textbf{46.4$\mathbf{_{0.5\uparrow}}$} & \textbf{71.8} & \textbf{48.3} & \textbf{56.7} & 53.4 & 50.0 & 52.1$\mathbf{_{0.1\downarrow}}$ & \textbf{75.8} \\
          & SCCAN (ICCV'23)~\cite{xu2023self} & 38.3 & 46.5 & 43.0 & 41.5 & 42.3 & 66.9 & 43.4 & 52.5 & 54.5 & 47.3 & 49.4 & 71.8 \\
          & SCCAN + \textbf{AENet} & 40.3 & \textbf{50.4} & 47.9 & 44.9 & 45.9$\mathbf{_{3.6\uparrow}}$ & 71.2 & 45.8 & 56.3 & \textbf{55.8} & \textbf{53.4} & \textbf{52.8$\mathbf{_{3.4\uparrow}}$} & 74.3 \\
    \midrule
    \multirow{13}[0]{*}{ResNet50} & PFENet (TPAMI'20) & 36.5  & 38.6  & 35.0  & 33.8  & 35.8  & -     & 36.5  & 43.3  & 38.0  & 38.4  & 39.0  & - \\
          & DACM (ECCV'22)~\cite{xiong2022doubly} & 37.5  & 44.3  & 40.6  & 40.1  & 40.6  & 68.9  & 44.6  & 52.0  & 49.2  & 46.4  & 48.1  & 71.6 \\
          & ABCNet (CVPR'23)~\cite{wang2023rethinking} & 42.3  & 46.2  & 46.0  & 42.0  & 44.1  & 69.9  & 45.5  & 51.7  & 52.6  & 46.4  & 49.1  & 72.7 \\
          & FECANet (TMM'23)~\cite{liu2023fecanet} & 38.5  & 44.6  & 42.6  & 40.7  & 41.6  & 69.6  & 44.6  & 51.5  & 48.4  & 45.8  & 47.6  & 71.1 \\
          & RiFeNet (AAAI'24)~\cite{bao2023relevant} & 39.1  & 47.2  & 44.6  & 45.4  & 44.1  & -     & 44.3  & 52.4  & 49.3  & 48.4  & 48.6  & - \\
          & BAM (CVPR'22)~\cite{lang2022learning} & 43.4  & 50.6  & 47.5  & 43.4  & 46.2  & -     & 49.3  & 54.2  & 51.6  & 49.6  & 51.2  & - \\
          & MIANet (CVPR'23)~\cite{yang2023mianet} & 42.5  & 53.0  & 47.8  & 47.4  & 47.7  & 71.5  & 45.8  & 58.2  & 51.3  & 51.9  & 51.7  & 73.1 \\
          \cmidrule{2-14}
          & CyCTR (NIPS'21)~\cite{zhang2021few} & 38.9 & 43.0 & 39.6 & 39.8 & 40.3 & - & 41.1 & 48.9 & 45.2 & 47.0 & 45.6 & - \\
          & CyCTR + \textbf{AENet} & 42.2 & 53.1 & 47.4 & 45.3 & 47.0$\mathbf{_{6.7\uparrow}}$ & 71.8 & 49.0 & 59.0 & 50.6 & 51.1 & 52.4$\mathbf{_{6.8\uparrow}}$ & 75.0 \\
          & SCCAN (ICCV'23)~\cite{xu2023self} & 40.4 & 49.7 & 49.6 & 45.6 & 46.3 & 69.9 & 47.2 & 57.2 & \textbf{59.2} & 52.1 & 53.9 & 74.2 \\
          & SCCAN + \textbf{AENet} & 43.1 & 56.0 & 50.3 & 48.4 & 49.4$\mathbf{_{3.1\uparrow}}$ & 73.6 & 51.7 & 61.9 & 57.9 & 55.3 & 56.7$\mathbf{_{2.8\uparrow}}$ & 76.5 \\
          & HDMNet$^*$ (CVPR'23)~\cite{peng2023hierarchical} & 44.8 & 54.9 & 50.0 & 48.7 & 49.6 & 72.1 & 50.9 & 60.2 & 55.0 & 55.3 & 55.3 & 74.9 \\
          & HDMNet + \textbf{AENet} & \textbf{45.4} & \textbf{57.1} & \textbf{52.6} & \textbf{50.0} & \textbf{51.3$\mathbf{_{1.7\uparrow}}$} & \textbf{74.4} & \textbf{52.7} & \textbf{62.6} & 56.8 & \textbf{56.1} & \textbf{57.1$\mathbf{_{1.8\uparrow}}$} & \textbf{78.5} \\
          \bottomrule
    \end{tabular}
  }
  % \vspace{-1em}
\end{table}

To well validate the effectiveness of
% the proposed methodology,
{\cam AENet,}
we plug
% AENet
{\cam it}
into
% CyCTR~\cite{zhang2021few} and SCCAN~\cite{xu2023self},
{\cam CyCTR~\cite{zhang2021few}, SCCAN~\cite{xu2023self} and HDMNet~\cite{peng2023hierarchical},}
and conduct experiments under both 1-shot and 5-shot settings, with VGG16~\cite{simonyan2014very} and ResNet50~\cite{he2016deep} as pretrained backbones.

\noindent
\textbf{PASCAL-5$^i$.}
The quantitative results on PASCAL-5$^i$ are shown in \cref{tab:quantitative_pascal}, we could observe that our AENet plug-in consistently helps to improve the
% two 
{\cam selected}
baselines by large margins. For example, by taking ResNet50 as the pretrained backbone, the mean mIoU score of SCCAN can be boosted from 66.8\% to 69.8\% on PASCAL-5$^i$ under 1-shot setting, and the improvement can be increased to 3.8\% when 5 support pairs are provided. Similarly, FB-IoU can be improved by 2.3\% and 2.7\%, respectively. Notably, the 1-shot performance gain on CyCTR can be as high as 4.8\%, showing the superiority of AENet.
Moreover, SCCAN does not behave as well as the best baseline HDMNet, but the incorporation of AENet can help it to outperform HDMNet by considerable margins, \eg, 66.6\% vs. 65.1\% (VGG16, 1-shot) and 74.1\% vs. 71.8\% (ResNet50, 5-shot).
{\cam Besides, HDMNet can be improved by 2.4\% (ResNet50, 5-shot) with AENet.}

\noindent
\textbf{COCO-20$^i$.}
COCO-20$^i$ appears to be a more challenging dataset, as the images usually contain small objects, multiple objects, and their BG are quite complex. Unfortunately, such characteristics would make the aforementioned issues much more severe, \eg, with the same receptive field, small FG objects are more likely to be mingled with more BG information, compared with large FG objects.
The results are displayed in \cref{tab:quantitative_coco}, and note that the results of HDMNet are reproduced by us, because their validation data lists of COCO-20$^i$ are different from the uniformly deployed ones in other baselines. Similar to PASCAL-5$^i$, our AENet can well boost the performance of
% CyCTR and SCCAN.
{\cam CyCTR, SCCAN and HDMNet.}
In particular, AENet can help to improve the baseline better on COCO-20$^i$ than on PASCAL-5$^i$, \eg, 3.6\% vs. 1.9\% with SCCAN (VGG16, 1-shot), and 6.7\% vs. 4.8\% with CyCTR (Res50, 1-shot). We contribute it to the fact that the FG-FG matching in baselines become less effective in COCO-20$^i$, while the proposed AENet can effectively mitigate this issue.

\begin{figure}[tb]
  \centering
  \includegraphics[width=1\linewidth]{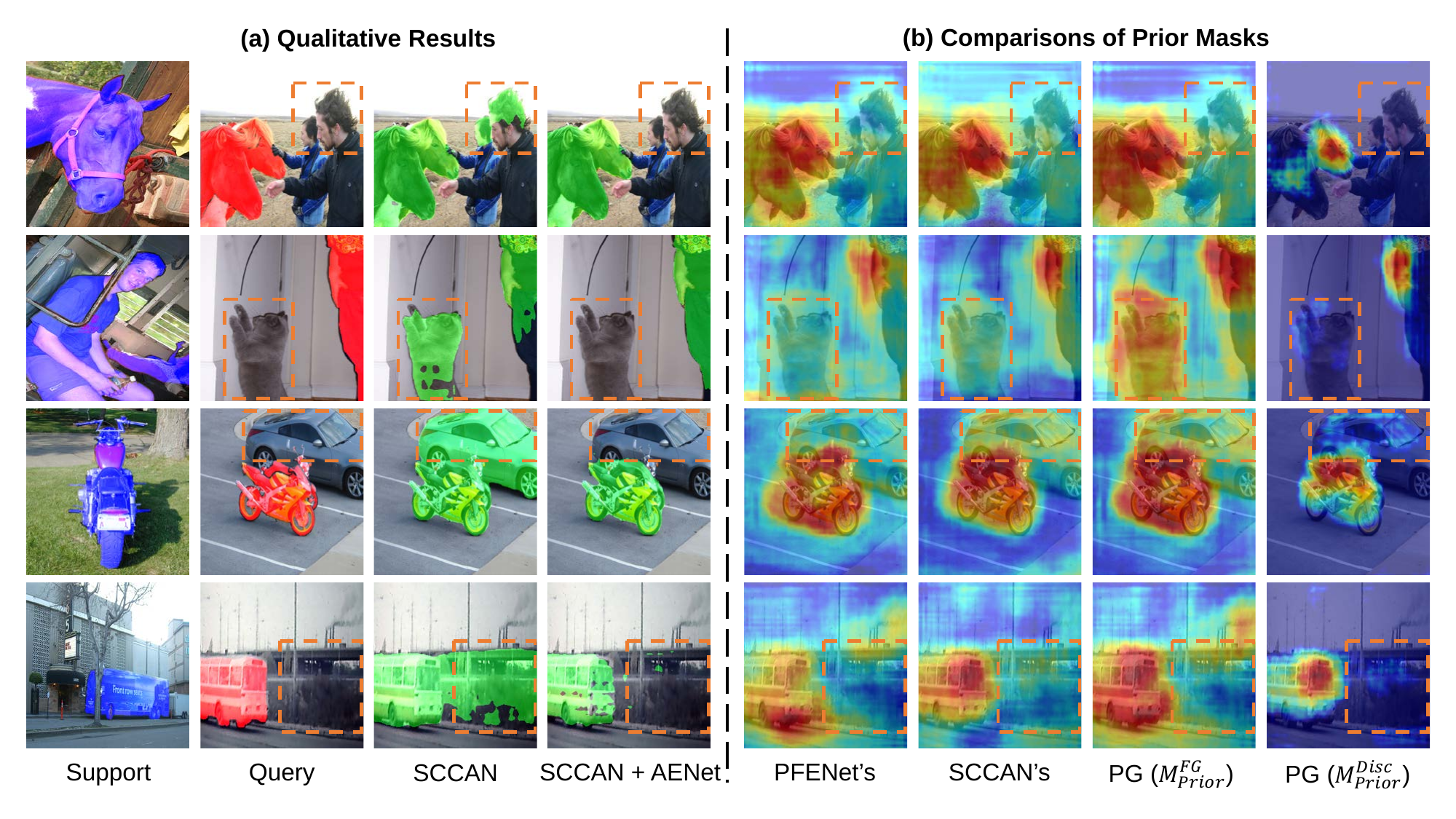}
  \caption{Illustration of (a) qualitative results and (b) different training-agnostic prior masks. In (b), the proposed PG is compared with the existing prior masks from PFENet~\cite{tian2020prior} and SCCAN~\cite{xu2023self}. We use some rectangles (in \textcolor{orange}{orange}) to highlight some challenging areas, where existing prior masks wrongly activate them as FG and lead to wrong predictions, but our PG ($M_{Prior}^{Disc}$) can suppress them well.
  }
  \label{fig:qualitative_prior}
  % \vspace{-1em}
\end{figure}

\subsection{Ablation Study}
\label{sec:ablation_study}

In this section, extensive ablation studies are conducted to validate the effectiveness of AENet. Unless explicitly specified, the experiments are conducted on PASCAL-5$^i$ with ResNet50 as the pretrained backbone, under 1-shot setting.

\noindent
\textbf{Qualitative results and prior masks comparisons.}
To have a clearer understanding of the proposed modules to the final predictions, we jointly depict some qualitative results, as well as the visual comparisons among different prior masks in \cref{fig:qualitative_prior}.
Specifically, the proposed PG is compared with two existing prior mask generation methods from PFENet~\cite{tian2020prior} and SCCAN~\cite{xu2023self}.
Although they can roughly locate the query FG, they merely measure the similarities between each query pixel and all the support FG pixels to determine if the current query pixel is more likely to be FG or BG, regardless of the fact that the FG/BG features are mingled with BG/FG features (as explained in \cref{sec:introduction}). In this way, some query BG pixels also contain query FG features, so there exist some similarities and they get wrongly activated.
Instead, our PG not only measures the similarities with support FG ($M_{Prior}^{FG}$), but also with support BG ($M_{Prior}^{BG}$), then perform a subtraction operation to obtain the discriminative prior mask $M_{Prior}^{Disc}$. As shown in \cref{fig:qualitative_prior}(b), $M_{Prior}^{FG}$ suffer from the same problem as PFENet and SCCAN, but $M_{Prior}^{Disc}$ can suppress the wrongly activated areas well. As a result, our plug-in AENet helps to make more accurate predictions.

\begin{table}[t]
    \begin{minipage}{0.45\linewidth}
        \caption{Component-wise ablation study with SCCAN.}
        \label{tab:component_ablation}
        \centering
        \resizebox{1\linewidth}{!}{
            \begin{tabular}{@{}cccccccc@{}}
            \toprule
            PG & AE & BAM & 5$^0$ & 5$^1$ & 5$^2$ & 5$^3$ & Mean \\
            \midrule
              &  &  & 68.3 & 72.5 & 66.8 & 59.8 & 66.8  \\
            \checkmark &  &  & 69.7 & 73.1 & 66.9 & 61.5 & 67.8$\mathbf{_{1.0\uparrow}}$ \\
              & \checkmark &  & 69.3 & 72.8 & 67.4 & 62.1 & 67.9$\mathbf{_{1.1\uparrow}}$ \\
            \checkmark & \checkmark &  & 70.1 & 73.4 & 66.9 & 63.1 & 68.3$\mathbf{_{1.5\uparrow}}$ \\
            \midrule
            \checkmark &  & \checkmark & 71.4 & 75.1 & 68.0 & 62.4 & 69.2$\mathbf{_{2.4\uparrow}}$ \\
              & \checkmark & \checkmark & 70.7 & 74.9 & 67.7 & 62.9 & 69.1$\mathbf{_{2.3\uparrow}}$ \\
            \checkmark & \checkmark & \checkmark & \textbf{72.2} & \textbf{75.5} & \textbf{68.5} & \textbf{63.1} & \textbf{69.8$\mathbf{_{3.0\uparrow}}$} \\
            \bottomrule
            \end{tabular}
        }
    \end{minipage}
    \hfill
    \begin{minipage}{0.5\linewidth}
        \caption{Detailed ablation study on PG and AE.}
        \label{tab:detailed_ablation_dpm_ae}
        \centering
        \resizebox{1\linewidth}{!}{
            \begin{tabular}{@{}cc|c|ccccc@{}}
            \toprule
            \multicolumn{2}{c|}{PG} & AE & \multirow{2}{*}{5$^0$} & \multirow{2}{*}{5$^1$} & \multirow{2}{*}{5$^2$} & \multirow{2}{*}{5$^3$} & \multirow{2}{*}{Mean} \\
            $M_{Prior}^{Disc}$ & $M_{Prior}^{FG}$ & \cref{eq:mid_dpm} & & & & & \\
            \midrule
              &  &  & 68.3 & 72.5 & 66.8 & 59.8 & 66.8  \\
            \midrule
            \checkmark &  &  & 70.7 & 73.2 & 65.7 & 60.4 & 67.5 \\
            \checkmark & \checkmark &  & 69.7 & 73.1 & 66.9 & 61.5 & 67.8 \\
            \midrule
             &  & $M^{FG}$ & 69.0 & 72.2 & 66.5 & 59.8 & 66.9 \\
             &  & $M^{Disc}$ & 69.3 & 72.8 & 67.4 & 62.1 & 67.9 \\
            \bottomrule
            \end{tabular}
        }
    \end{minipage}
    % \vspace{-1em}
\end{table}

\noindent
\textbf{Component-wise ablation study.}\label{sec:component_ablation}
BAM~\cite{lang2022learning} is a controversial baseline that extends the standard FSS setting by:
(1) They wrap the pretrained backbone with PSPNet~\cite{zhao2017pyramid}, and fine-tune it for segmenting the base classes, which is more appropriate for FSS;
(2) Then, the base class predictions are obtained and suppressed as BG regions during testing, which improves the accuracy.
Despite of the controversy, most of the latest baselines~\cite{peng2023hierarchical,wang2023adaptive,yang2023mianet,lang2023base} follow BAM's setting.
Hence, we experiment with both settings for better comparisons.
It could be observed from \cref{tab:component_ablation} that the mean mIoU of SCCAN starts with 66.8\%, and the score can be boosted to 67.8\% and 67.9\% after integrating PG and AE module, respectively. Then, when both PG and AE are utilized, the score is further increased to 68.3\% (+1.5\%), showing the effectiveness of the proposed AENet plug-in.
With BAM's ensemble, the final score can be as high as 69.8\%.
% which surpasses the best baseline HDMNet~\cite{peng2023hierarchical}.

\noindent
\textbf{Further discussion on PG.}
As described in \cref{sec:dpm}, we generate three prior masks, including $M_{Prior}^{FG}$ (measured with support FG features), $M_{Prior}^{BG}$ (measured with support BG features), and $M_{Prior}^{Disc}$ (clipped subtraction).
Among them, $M_{Prior}^{FG}$ is visually similar to the existing prior masks, while $M_{Prior}^{Disc}$ appears to have better quality (as displayed in \cref{fig:qualitative_prior}(b)).
It can be observed from \cref{tab:detailed_ablation_dpm_ae}, when we replace SCCAN's prior mask with $M_{Prior}^{Disc}$, the mean mIoU score can be boosted by 0.7\%. If we further use $M_{Prior}^{FG}$, the final improvement can be 1.0\%. We contribute the improvement to the different functions of these masks, \eg, (1) $M_{Prior}^{FG}$ is responsible for roughly locating the query FG regions. Although the generated prior masks also have high responses on BG regions, in most cases, the complete FG objects have already been included; and (2) for $M_{Prior}^{Disc}$, it only activates the most discriminative query FG regions that are dissimilar to the support BG features. Therefore, when they are both utilized, the model would firstly be aware of the regions it should focus on, then it can take the discriminative regions as anchor, and find other similar parts in the query image where there would be no intra-class differences~\cite{okazawa2022interclass,fan2022self}.

\noindent
\textbf{Detailed ablation on AE.}
We further conduct experiments to prove that the ``subtraction'' operation is critically important. As illustrated in the last two rows of \cref{tab:detailed_ablation_dpm_ae}, we use $M^{Disc}$ and $M^{FG}$ to distinguish the cases where there is a subtraction operation (\cref{eq:disc_prior}) or not.
According to the table, we can observe that merely taking support FG information into consideration can hardly make any improvement (66.8\% vs. 66.9\%).
Instead, if $M^{Disc}$ is alternatively utilized, the improvement can be as high as 1.1\%.
Then, we explain this phenomenon as follows:
(1) The query FG pixels' features are doped with \emph{base class-specific} BG features, because different classes tend to have their own sets of BG classes. Therefore, when the auxiliary loss (\cref{eq:aux_loss}) is applied for regularization, the model would be likely to learn \emph{base class-specific} operations to detach the mingled BG features from query FG features. Thus, the model gets biased, and the learned pattern cannot be safely applied to the novel classes;
(2) Instead, the subtraction operation (\cref{eq:disc_prior}) can provide the model with some \emph{class-agnostic} guidance, so the learned pattern can be relatively uniform for all the classes.

\begin{figure}[tb]
    \centering
    \includegraphics[width=1\linewidth]{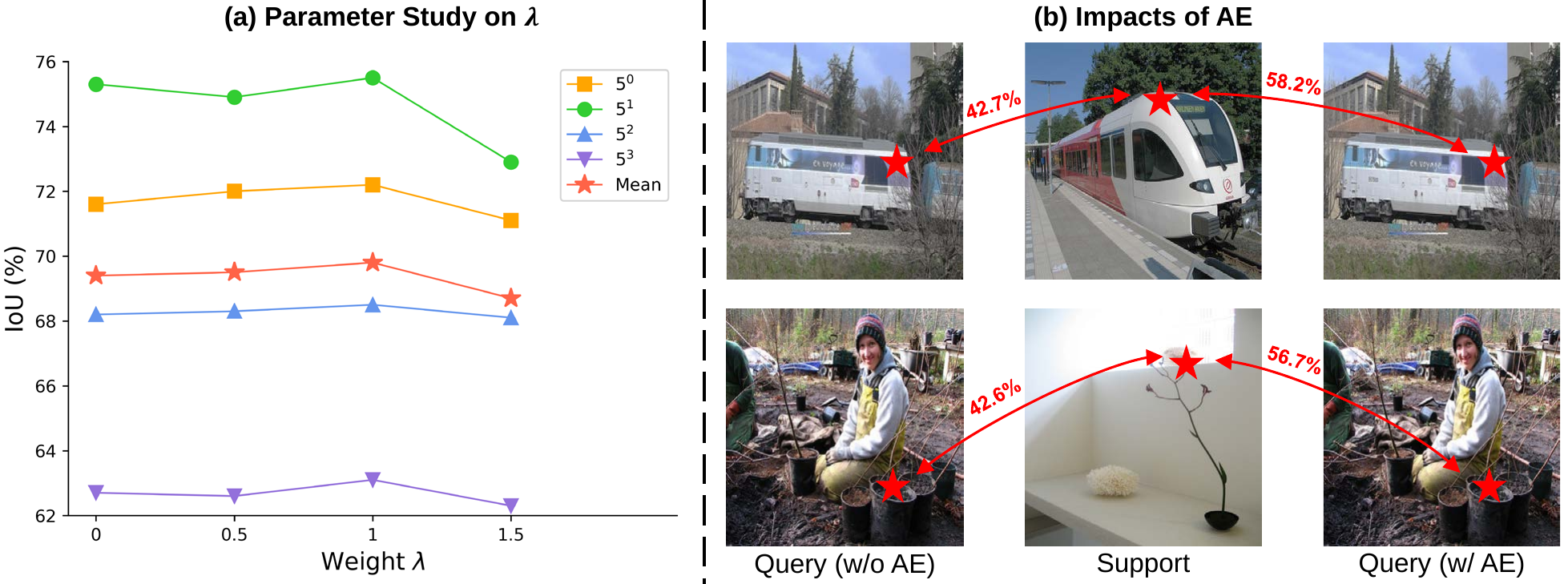}
    \caption{Illustration of (a) parameter study on weight $\lambda$, and (b) the impacts of AE. In (b), \textcolor{red}{$\star$} represent FG pixels, the values are measured by cosine similarity.
    }
    \label{fig:weight_lambda_ae_viz}
    % \vspace{-1em}
\end{figure}

\noindent
\textbf{Parameter study on loss weight $\lambda$.}\label{sec:weight_lambda}
We set the loss weight $\lambda$ (in \cref{eq:overall_loss}) as \{0, 0.5, 1, 1.5\} to explore their effects, and show the results in \cref{fig:weight_lambda_ae_viz}(a). It could be observed:
(1) Even if we do not apply additional supervision signals on the generated discriminative mask in AE, the mean mIoU score can already reach 69\%+, showing the effectiveness of the discriminative mask;
(2) When $\lambda=1$, the best performance is achieved;
(3) When $\lambda>1$, the performance starts decreasing.
Therefore, we set $\lambda=1$ by default in the paper.

\noindent
\textbf{Impacts of AE.}
% Recall that AE is proposed to mitigate the \emph{feature ambiguity} issue, so as to improve the FG-FG matching in cross attention.
{\cam AE is designed to mitigate the \emph{feature ambiguity} issue, so as to improve the FG-FG matching.}
To show the impacts of AE, we draw two examples in \cref{fig:weight_lambda_ae_viz}(b), where the features before and after the AE module are taken to measure the query-support similarity. As shown in the figure, AE module consistently helps to improve the FG-FG similarity between query and support FG pixels. In this way, query FG pixels can fuse more FG features from the support samples, \ie, the support information is well utilized for more effective FSS.

% \noindent
% \textbf{Study on weak support annotations.}
% TODO

% ========================================
% Conclusion
% ========================================
\section{Conclusion}
\label{sec:conclusion}

In this paper, we identify the negative impacts of feature ambiguity to the cross attention modules in FSS. To alleviate the issue, we design a plug-in ambiguity elimination network (AENet) which includes a prior generator (PG) and an ambiguity eliminator (AE) module. Learning-agnostic PG is responsible for roughly locating the query FG objects, and highlighting the most discriminative query FG regions. AE utilizes the discriminative query FG features to rectify the features, so as to enhance the cross attentions. We plug AENet to
% two
{\cam three}
FSS baselines, and can improve their performance by large margins. Extensive experiments have been conducted to validate the effectiveness of AENet.

\smallskip
\noindent
\textbf{Acknowledgement.}
This study is supported under the RIE2020 Industry Alignment Fund - Industry Collaboration Projects (IAF-ICP) Funding Initiative, as well as cash and in-kind contribution from the industry partner(s).

% ---- Bibliography ----
%
% BibTeX users should specify bibliography style 'splncs04'.
% References will then be sorted and formatted in the correct style.
%
\bibliographystyle{splncs04}
\bibliography{egbib}
\end{document}

% ---------------------------------------------------------------
% TODO REVIEW: Replace with your title
\title{Eliminating Feature Ambiguity for \\ Few-Shot Segmentation} 

% TODO REVIEW: If the paper title is too long for the running head, you can set
% an abbreviated paper title here. If not, comment out.
% \titlerunning{DiscFSS}

% TODO FINAL: Replace with your author list. 
% Include the authors' OCRID for the camera-ready version, if at all possible.
\author{Qianxiong Xu\inst{1}\orcidlink{0000-0001-9175-6783} \and
Guosheng Lin\inst{1}\orcidlink{0000-0002-0329-7458} \and
Chen Change Loy\inst{1}\orcidlink{0000-0001-5345-1591} \and
Cheng Long\inst{1}\orcidlink{0000-0001-6806-8405} \and \\
Ziyue Li\inst{2}\orcidlink{0000-0003-4983-9352} \and
Rui Zhao\inst{3}
}
% TODO FINAL: Replace with an abbreviated list of authors.
\authorrunning{Q. Xu et al.}
% First names are abbreviated in the running head.
% If there are more than two authors, 'et al.' is used.

% TODO FINAL: Replace with your institution list.
\institute{S-Lab, Nanyang Technological University \and University of Cologne \and SenseTime Research\\
\email{\{qianxiong.xu, gslin, ccloy, c.long\}@ntu.edu.sg},\\
\email{zlibn@wiso.uni-koeln.de},
\email{zhaorui@sensetime.com}}

% \maketitle

% ========================================
% Supplementary
% ========================================
\appendix

% \section{Code}
% \label{sec:code}

% We upload the source code in the attached file. Besides, we include some instructions for obtaining the employed PASCAL-5$^i$ and COCO-20$^i$ datasets, as well as some anonymous google drive links for our pretrained models.

\section{Parameter Amount and FLOPs}
\label{sec:param_amount_flops}

\begin{table}[h]
  \caption{Impacts of AENet on parameters and FLOPs.}
  \label{tab:param_amount_flops}
  \centering
  \scalebox{.85}{
    \begin{tabular}{@{}l|cc@{}}
    \toprule
    Method & \#Parameters & \#FLOPs \\
    \midrule
    SCCAN & 35.0M & 480.9G \\
    SCCAN + AENet & 40.7M$_{14.0\%\uparrow}$ & 528.4G$_{9.0\%\uparrow}$ \\
    \bottomrule
    \end{tabular}
  }
\end{table}

Our proposed AENet serves as a plugin for existing cross attention-based few-shot segmentation (FSS) methods. Take SCCAN as an example, it has 8 self-calibrated cross attention (SCCA) blocks, thus we insert 8 ambiguity eliminators (AE) in total, with each of them inserted before 1 SCCA block. Besides, we replace its pseudo mask aggregation (PMA) module with our prior generator (PG).
We summarize the parameter amount, as well as the FLOPs, of SCCAN and SCCAN + AENet in \cref{tab:param_amount_flops}, and could observe that our proposed AENet is lightweight in both parameters and computations, \eg, the there is only a 9\% increase in terms of FLOPs.

\section{Weak Support Labels}
\label{sec:weak_support_labels}

\begin{table}[h]
  \caption{Performance comparisons with weak support labels (bounding boxes). The backbone is ResNet50. $^*$ show the performance with accurate pixel-wise labels.}
  \label{tab:weak_support_labels}
  \centering
  \scalebox{.85}{
    \begin{tabular}{@{}l|cccccc@{}}
    \toprule
    \multirow{2}[0]{*}{Method} & \multicolumn{6}{c}{1-shot} \\
    & 5$^0$ & 5$^1$ & 5$^2$ & 5$^3$ & Mean  & FB-IoU \\
    \midrule
    PANet & - & - & - & - & 45.1 & - \\
    CANet & - & - & - & - & 52.0 & - \\
    DPCN & 59.8 & 70.5 & 63.2 & 55.5 & 62.3 & - \\
    \midrule
    SCCAN & 67.3 & 71.8 & 65.6 & 58.0 & 65.7 & 75.5 \\
    SCCAN$^*$ & 68.3 & 72.5 & 66.8 & 59.8 & 66.8 & 77.7 \\
    \midrule
    SCCAN + AENet & 71.8 & 74.6 & 67.1 & 61.7 & 68.8 & 80.5 \\
    SCCAN + AENet$^*$ & 72.2 & 75.5 & 68.5 & 63.1 & 69.8 & 80.8 \\
    \bottomrule
    \end{tabular}
  }
\end{table}

Given a specific class, semantic segmentation relies on large number of manually annotated samples to learn its representative features for segmentation. The most inspiring thing of few-shot segmentation (FSS) is it can greatly reduce the annotation cost from more than thousands of samples to only 1 or 5 samples.
However, some existing studies think that even 1 or 5 pixel-wise labels still cost much, they further conduct experiments under the scenario where cheaper weak support labels are provided, \eg, bounding boxes.
It can be observed from \cref{tab:weak_support_labels} that the proposed AENet also works well with weak support labels, validating the effectiveness of our idea, \ie, mining discriminative query regions with a ``subtraction'' operation, which can mitigate the side-effects of the BG features mingled in FG features.

\section{More testing episodes on COCO-20$^i$}
\label{sec:more_test_coco}

\begin{table}[t]
  \caption{Testing results of 20,000 episodes on COCO-20$^i$ in terms of mIoU and FB-IoU. ``20$^i$'' shows the mIoU scores of 20 novel classes in fold $i$, ``Mean'' is the averaged mIoU score from all folds. $^*$ means testing with 4,000 episodes, and $^{\dagger}$ means testing with 20,000 episodes.}
  \label{tab:more_test_coco}
  \centering
  \resizebox{1\linewidth}{!}{
    \begin{tabular}{@{}cl|cccccc|cccccc@{}}
    \toprule
    \multirow{2}[0]{*}{Backbone} & \multirow{2}[0]{*}{Method} & \multicolumn{6}{c|}{1-shot}                    & \multicolumn{6}{c}{5-shot} \\
          &       & 5$^0$ & 5$^1$ & 5$^2$ & 5$^3$ & Mean  & FB-IoU & 5$^0$ & 5$^1$ & 5$^2$ & 5$^3$ & Mean  & FB-IoU \\
    \midrule
    \multirow{2}[0]{*}{VGG16} & SCCAN + AENet$^*$ & 40.3 & 50.4 & 47.9 & 44.9 & 45.9 & 71.2 & 45.8 & 56.3 & 55.8 & 53.4 & 52.8 & 74.3 \\
          & SCCAN + AENet$^{\dagger}$ & 39.4 & 49.9 & 46.2 & 44.9 & 45.1 & 71.1 & 45.2 & 56.0 & 55.0 & 52.6 & 52.2 & 74.3 \\
    \midrule
    \multirow{2}[0]{*}{ResNet50} & SCCAN + AENet$^*$ & 43.1 & 56.0 & 50.3 & 48.4 & 49.4 & 73.6 & 51.7 & 61.9 & 57.9 & 55.3 & 56.7 & 76.5 \\
          & SCCAN + AENet$^{\dagger}$ & 42.6 & 56.3 & 48.8 & 48.6 & 49.1 & 73.5 & 49.5 & 61.8 & 56.5 & 55.6 & 55.8 & 76.6 \\
          \bottomrule
    \end{tabular}
  }
\end{table}

Compared to PASCAL-5$^i$, COCO-20$^i$ contains much more images. Therefore, we follow PFENet to test SCCAN + AENet with 20,000 testing episodes, so as to obtain more robust results on COCO-20$^i$. The results are shown in \cref{tab:more_test_coco}, and it could be observed that there is no prominent performance drop, which means the proposed method is stable.

{\cam

\begin{table}[h]
  \centering
  \caption{Error bars evaluation on COCO-20$^i$. The random seeds are taken from \{0, 1, 2, 3, 321\} to generate 4,000 testing episodes. \textbf{Bold} values denote the best cases.}
  \label{tab:error_bars_evaluation}
  \scalebox{.85}{
    \begin{tabular}{@{}c|c|cccccc@{}}
    \toprule
    \multirow{2}[0]{*}{Method} & \multirow{2}[0]{*}{Seed} & \multicolumn{6}{c}{1-shot} \\
          &       & 20$^0$ & 20$^1$ & 20$^2$ & 20$^3$ & Mean  & FB-IoU \\
    \midrule
    \multirow{7}[0]{*}{HDMNet} & 0     & 45.5  & 55.3  & 49.6  & 46.7  & 49.3  & 71.9 \\
          & 1     & 45.3  & 54.9  & 50.8  & 48.3  & 49.8  & 71.8 \\
          & 2     & 44.9  & 54.2  & 50.0  & 48.7  & 49.5  & 72.2 \\
          & 3     & 44.1  & 54.9  & 51.9  & 48.6  & 49.9  & 72.1 \\
          & 321   & 44.8  & 54.9  & 50.0  & 48.7  & 49.6  & 72.0 \\
          \cmidrule{2-8}
          & Mean  & 44.9  & 54.9  & 50.5  & 48.2  & 49.6  & 72.0 \\
          & Std   & \textbf{0.6}   & 0.4   & 0.9   & 0.9   & 0.3   & 0.2 \\
    \midrule
    \multirow{7}[0]{*}{HDMNet + AENet} & 0     & 46.7  & 57.7  & 52.0  & 49.5  & 51.5  & 74.3 \\
          & 1     & 47.2  & 57.1  & 51.5  & 51.1  & 51.7  & 74.4 \\
          & 2     & 45.7  & 57.6  & 52.0  & 49.1  & 51.1  & 74.4 \\
          & 3     & 47.1  & 57.9  & 52.8  & 49.7  & 51.9  & 74.4 \\
          & 321   & 45.4  & 57.1  & 52.6  & 50.0  & 51.3  & 74.4 \\
          \cmidrule{2-8}
          & Mean  & \textbf{46.4}  & \textbf{57.5}  & \textbf{52.2}  & \textbf{49.9}  & \textbf{51.5}  & \textbf{74.4} \\
          & Std   & 0.8   & \textbf{0.4}   & \textbf{0.5}   & \textbf{0.7}   & \textbf{0.3}   & \textbf{0.1} \\
    \bottomrule
    \end{tabular}
  }
\end{table}

\section{Error Bars Evaluation}
\label{sec:error_bars_evaluation}

We further report the 1-shot error bars evaluation of HDMNet and HDMNet + AENet in \cref{tab:error_bars_evaluation}. It can be observed that with our plug-in AENet, (1) HDMNet can consistently be improved by considerable margins, and (2) the results are stable, as the standard deviation of mean mIoU score is only 0.3.
}

\section{More Qualitative Results}
\label{sec:more_qualitative}

\begin{figure}[h]
  \centering
  \includegraphics[width=1\linewidth]{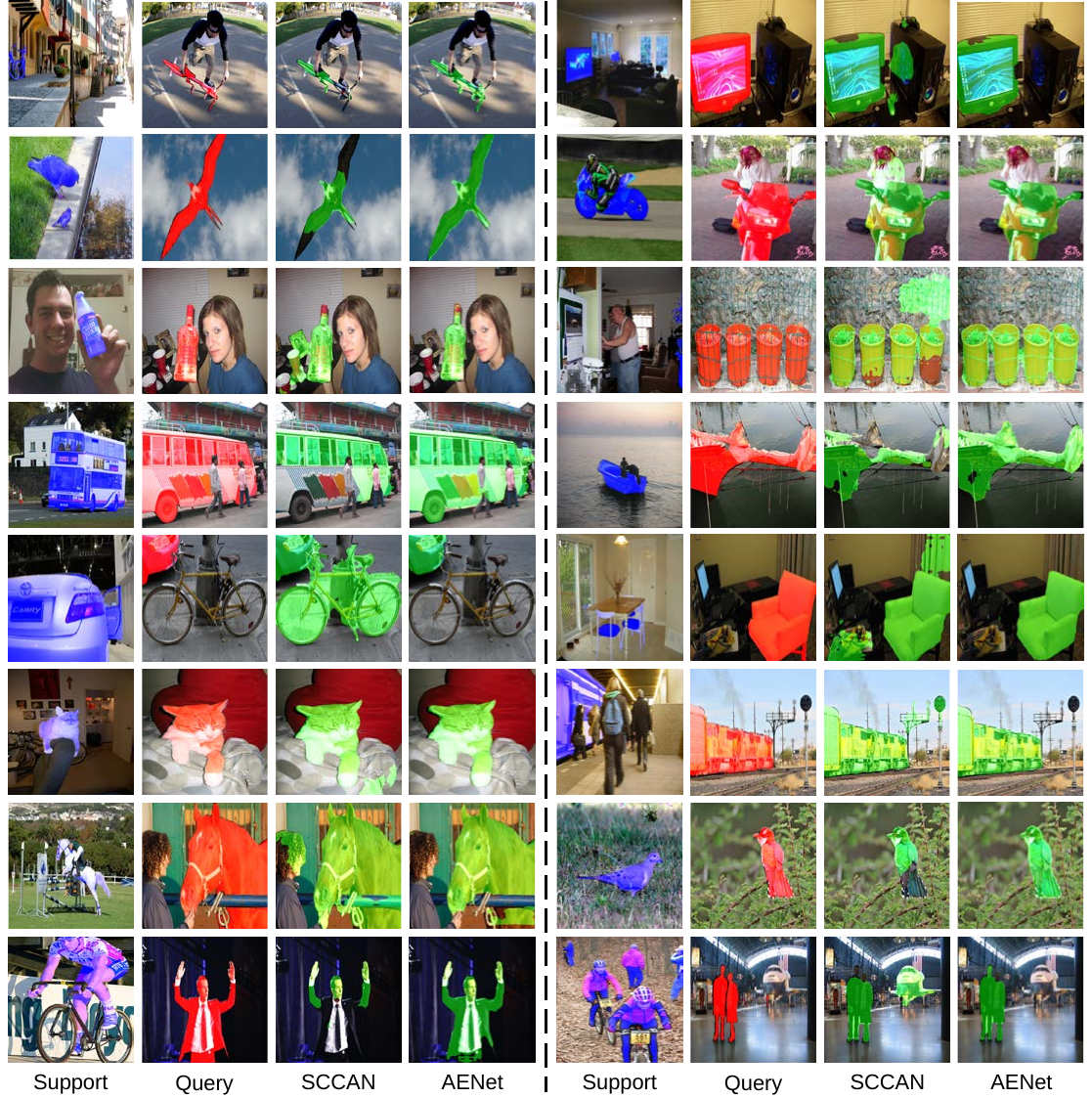}
  \caption{More qualitative results of SCCAN and SCCAN + AENet.
  }
  \label{fig:more_qualitative}
\end{figure}

More qualitative results of SCCAN and SCCAN + AENet are displayed in \cref{fig:more_qualitative}, where we can observe a stable improvement of AENet over SCCAN.

\section{More Visualizations of Prior Masks}
\label{sec:more_vis_prior}

\begin{figure}[h]
  \centering
  \includegraphics[width=0.9\linewidth]{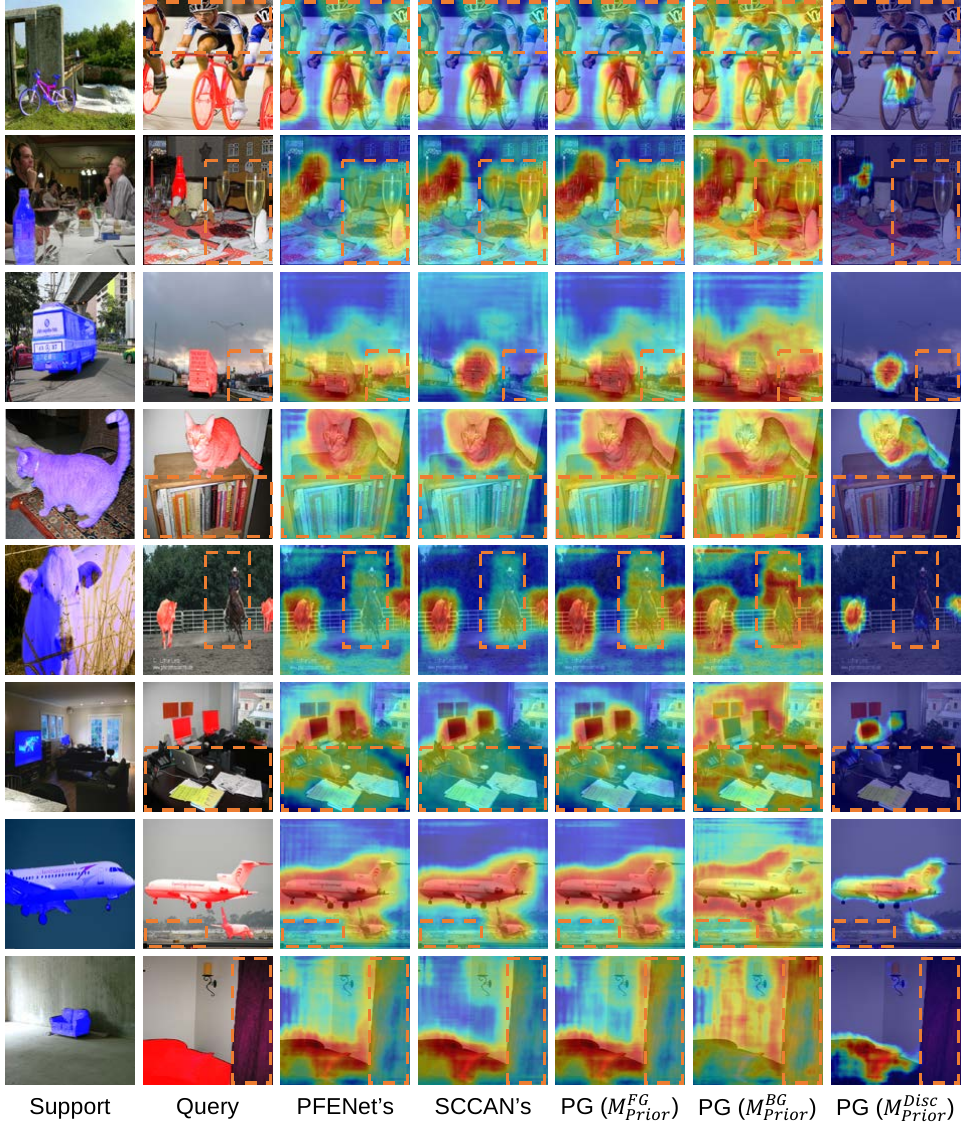}
  \caption{More visualizations results of different prior masks, including the prior masks from PFENet, SCCAN and our prior generator (PG). We use some \textcolor{orange}{orange} rectangles to highlight some challenging areas.
  }
  \label{fig:more_prior}
\end{figure}

The visualizations of discriminative prior masks serve as the direct evidences to the effectiveness of our main idea. In this section, we depict more examples in \cref{fig:more_prior}. We could observe that existing prior mask generation methods would mistakenly activate many wrong areas, and become ineffective, while our discriminative prior mask ($M_{Prior}^{Disc}$) can consistently suppress them well, which demonstrates the effectiveness of our ``subtraction'' operation in Eq. (4).

\section{Limitation and Future Direction}
\label{sec:limitation}

The main motivation of the designed plugin ambiguity elimination network (AENet) is to improve the query-support FG-FG matching for existing cross attention-based FSS methods. More concretely, this is achieved by mining discriminative query FG regions and then using them for query and support features refinement. Although the query and support FG pixels can consequently contain more FG information (so as to enhance FG-FG matching naturally), the query BG pixels would also be fused with the discriminative query FG features, making the refined query FG and BG features hard for separation. Therefore, a possible future direction is to design a module to prevent query BG pixels from fusing discriminative FG features.

% ---- Bibliography ----
%
% BibTeX users should specify bibliography style 'splncs04'.
% References will then be sorted and formatted in the correct style.
%
% \bibliographystyle{splncs04}
% \bibliography{egbib}